%% file: main.tex
\documentclass[twoside]{article}
\usepackage{primearxiv}
\usepackage{svg}
\usepackage{times}
\usepackage{soul}
\usepackage{url}
\usepackage[hidelinks]{hyperref}
\usepackage[utf8]{inputenc}
\usepackage[small]{caption}
\usepackage{graphicx}
\usepackage{amsmath}
\usepackage{amsthm}
\usepackage{booktabs}
\usepackage{algorithm}
\usepackage{algorithmic}
\usepackage[switch]{lineno}
\usepackage{macro}
\definecolor{darkgreen}{rgb}{0.0, 0.2, 0.13}

\title{Large Language Model Meets Constraint Propagation}

\author{
    Alexandre Bonlarron \\
    Université Côte d’Azur, CNRS, I3S, France \\
    alexandre.bonlarron@gmail.com \\
    \and
    Florian Régin \\
    Université Côte d'Azur, CNRS, I3S, France\\
    florian.regin06@gmail.com
    \and
    Elisabetta De Maria \\
    Université Côte d'Azur, CNRS, I3S, France\\
    elisabetta.de-maria@univ-cotedazur.fr
    \and
    Jean-Charles Régin \\
    Université Côte d’Azur, CNRS, I3S, France\\
    jean-charles.regin@univ-cotedazur.fr
}

\begin{document}

\maketitle

\begin{abstract}

Large Language Models (LLMs) excel at generating fluent text but struggle to enforce external constraints because they generate tokens sequentially without explicit control mechanisms. GenCP addresses this limitation by combining LLM predictions with Constraint Programming (CP) reasoning, formulating text generation as a Constraint Satisfaction Problem (CSP). In this paper, we improve GenCP by integrating Masked Language Models (MLMs) for domain generation, which allows bidirectional constraint propagation that leverages both past and future tokens. This integration bridges the gap between token-level prediction and structured constraint enforcement, leading to more reliable and constraint-aware text generation. Our evaluation on COLLIE benchmarks demonstrates that incorporating domain preview via MLM calls significantly improves GenCP's performance. Although this approach incurs additional MLM calls and, in some cases, increased backtracking, the overall effect is a more efficient use of LLM inferences and an enhanced ability to generate feasible and meaningful solutions, particularly in tasks with strict content constraints.
\end{abstract}

\input{inc/intro2}

\input{inc/background2}

\input{inc/method2}

\input{inc/result}

\input{inc/discussion}

\input{inc/Conclusion}

\bibliographystyle{unsrt}
\bibliography{biblio}

\end{document}

%% file: inc/intro2.tex
\section{Introduction}

The landscape of Large Language Models (LLMs) is evolving rapidly, with new features and capabilities emerging near-weekly. Despite their impressive fluency and versatility, current LLMs often struggle to adhere to specific rules or regulations during text generation \cite{yao2024collie}. This limitation arises because LLMs are primarily designed as text predictors without built-in mechanisms to enforce constraints. This lack of guarantee has become an increasingly important, almost fundamental problem around the confident and safe usage of LLMs \cite{cardei2025constraineddiscretediffusion,geh2025adversarialtokenization}.
As a result, recent Natural Language Processing (NLP) trends have explored augmenting LLMs with external control mechanisms. Examples include Retrieval-Augmented Generation (RAG) (e.g., integrating search engine results), grounding techniques (e.g., injecting relevant documents into the context window), external knowledge bases, and reasoning frameworks such as chain-of-thought~\cite{Chains-of-though:2022} and tree-of-thought \cite{Yao-tot:2023}. However, these methods often provide only partial solutions to the broader problem of \emph{constrained text generation}—the task of generating text that meets mandatory requirements.

Structured methods from combinatorial optimization offer a promising alternative. For instance, heuristically-guided text generation utilizes techniques like Beam Search (BS) \cite{beamsearch2:2018,lu-etal-2022-neurologic} by exploring a subset of the search space. In contrast, Constraint Programming (CP) \cite{bonlarron-et-al:2023,bonlarron-regin:2024inter} has been used to formalizes the constrained generation task as a Constraint Satisfaction Problem (CSP), where:
\begin{itemize}
    \item Variables correspond to text words,
    \item Domains represent the allowed vocabulary,
    \item and Constraints define the admissibility of word sequences.
\end{itemize}
Once the feasible solution set is found through an exhaustive search, the LLM performs a curation phase to rank these candidates based on criteria such as fluency and coherence. While effective in scenarios dominated by strong constraints, these CP-based approaches require multiple, complex processing steps and are parameter-sensitive. They suffer from the combinatorial explosion when constraints are weaker. This limitation arises from the loosely coupled two-step process, which separates CP and LLM, thereby hindering the direct integration of the LLM into the solving procedure. Ultimately, an LLM surrogate is employed to partially mitigate the combinatorial explosion, albeit while remaining sensitive to the cost of LLM inference in the curation phase ($>>$100\,ms) \cite{bonlarron-regin:2024markov}.

A recent and up-and-coming line of research leverages LLMs to guide CP-based text generation: (GenCP) \cite{regin-demaria-bonlarron:2024}, achieving 100\% constraint satisfaction on recent benchmarks \cite{yao2024collie} (sentence generation tasks). In this strongly coupled approach, the LLM first generates candidate tokens (using top-$p$ or top-$k$ sampling \cite{top-p-ICLR-2020}), which are then refined by a CP solver through a method that sequentially instantiates variables while maintaining local consistency in a backtracking search.

Unfortunately, the GenCP method is based on the modeling approach proposed by Bonlarron et al., which assumes a one-to-one correspondence between words and variables. However, this assumption does not hold in general for LLMs, as they process text at the token level, where tokens may correspond to entire words, subwords, or even individual characters, depending on the tokenizer \cite{geh2024signal}. 

This modeling choice influenced how GenCP managed variables, ultimately limiting its ability to generate high-quality and feasible outputs. The original GenCP paper acknowledged the need for a token-based approach rather than a word-based one. In this work, we extend GenCP by introducing meta-variables, enabling it to manage tokens more effectively and improve both constraint satisfaction and text generation quality.

However, relying solely on autoregressive LLMs introduces a significant limitation: these models generate text left to right and do not give information about future tokens. In principle, if a filtering algorithm (propagator) can detect inconsistencies faster than the search procedure would otherwise uncover them, then early propagation of the constraints is highly beneficial. In other words, effective propagation reduces the search space by pruning inconsistent candidates before the solver suffers the cost of further search exploration. Since standard CP solvers rely on propagation, where constraints between distant variables help refine candidate domains, this unidirectional generation restricts the solver’s ability to forecast future values (e.g., predicting the domain $D(X_8)$ solely on assignments $X_1$ to $X_5$). Consequently, the CP propagation mechanism becomes \emph{starved} of necessary domain values, leading to redundant computations and search \emph{thrashing} during backtracking.

\begin{figure}
    \centering
    \includegraphics[width=0.75\linewidth]{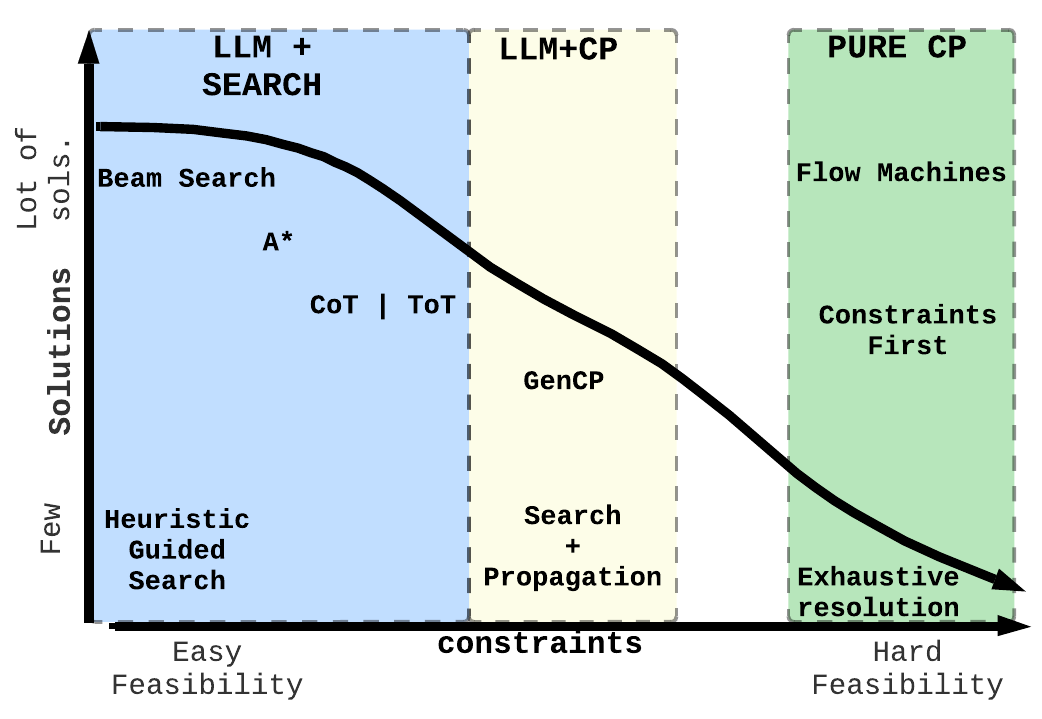}
    \caption{This diagram positions different methods along the x-axis based on feasibility difficulty for constrained text generation, from 'Easy Feasibility' to 'Hard Feasibility'. Methods such as Beam Search \protect\cite{beamsearch2:2018}, A* \protect\cite{lu-etal-2022-neurologic}, and Tree of Thoughts (ToT) \protect\cite{yao2024collie}, GenCP \protect\cite{regin-demaria-bonlarron:2024}, Pachet's ERC Flow Machines (Markov +X) \protect\cite{pachet-roy:2011}, and Constraints First \protect\cite{bonlarron-et-al:2023} illustrate this progression.}

    \label{fig:cpml}
\end{figure}

To overcome this limitation, we propose leveraging Masked Language Models (MLMs), such as BERT \cite{BERT}, to enhance constraint propagation. Unlike autoregressive models, MLMs process entire sequences bidirectionally by predicting masked words based on both preceding and following context. For example, given the prompt “The city of London is the \texttt{[MASK]} of the United Kingdom,” BERT correctly predicts \emph{capital}. This bidirectional capability naturally aligns with CP solvers \cite{bessiere2006constraint}, as it enables constraints to be propagated across the entire sequence rather than solely in a left-to-right manner. In practice, if tokens $X_1$ through $X_5$ are assigned, an MLM can generate relevant candidate tokens for $X_7$, thereby refining the assignment decision for $X_6$ and reducing the likelihood of dead-ends and unproductive search efforts. Ultimately, while the backtracking search is guided by an autoregressive LLM in a left-to-right fashion, any constraint that involves long-range dependencies prompts the solver to leverage an MLM to enrich its candidate domains, thereby enabling improved filtering and more informed assignment decisions.

This article makes the following contributions:
\begin{itemize}
    \item it introduces a novel integration of LLMs and CP that leverages bidirectional MLMs to enhance constraint propagation in text generation.
    \item it improves domain filtering by incorporating both past and future token assignments, thereby mitigating the limitations of autoregressive generation alone.
\end{itemize}

This article is organized as follows: Section~\ref{section:background} briefly introduces the necessary background on Constraint Programming and NLP techniques. Section~\ref{section:method} details our proposed approach. Section~\ref{section:result} presents experimental results, and Section~\ref{section:discussion} discusses limitations, open challenges, and future research directions.
Section~\ref{section:conclusion} concludes the paper.



%% file: inc/background2.tex
\section{Background}
\label{section:background}

This section briefly reviews the key concepts and related work underpinning our approach, focusing on constrained text generation, LLM decoding, and GenCP.

\subsection{Related Work in Constrained Generation}
Early work in CP demonstrated its potential in creative domains such as text and music generation. For instance, the ERC Flow Machines led by François Pachet explored CP for style modeling \cite{pachet-roy:2011}, lyric composition \cite{barbieri-et-al-lyrics:2012}, and many creative applications \cite{papadopoulos-roy-pachet:2014,papadopoulos-roy-etal:15,papadopoulos-roy-pachet:2016assisted}. More recent efforts have applied CP to standardized sentence generation \cite{bonlarron-et-al:2023,bonlarron-regin:2024markov,bonlarron-regin:2024inter} and even advanced musical composition \cite{sprockeels-vanroy:2024}. These contributions underscore CP's versatility and relevance to express complex requirements through constraints or exhaustively explore search space in generative setup.

\subsection{Decoding Strategies in LLMs}
Autoregressive LLMs generate text by predicting tokens sequentially. Common decoding strategies include:
\begin{itemize}
    \item \textbf{Greedy Decoding:} Selects the most probable token at each time step but may miss globally optimal sequences.
    \item \textbf{Stochastic Sampling:} Top-$k$ and top-$p$ (nucleus) sampling \cite{top-p-ICLR-2020} introduce diversity by sampling from a subset of high-probability tokens.
    \item \textbf{Beam Search:} Maintains the $k$ most likely sequences at each step \cite{hokamp-liu:2017,beamsearch2:2018}. Although beam search can improve overall results, it relies on local token probabilities and is tailor-made to manage local constraints such as keyword constraints. It was extended to deal with constraints as logic predicates in conjunctive normal form \cite{lu-etal-2021-neurologic}.
    \item \textbf{Beam Search + A*} Beam search augmented with an A* look-ahead \cite{lu-etal-2022-neurologic} is employed to enhance overall sequence probabilities in both unconstrained and constrained scenarios. This approach retains the top $k$ candidate partial solutions that are closest to satisfying the constraints. In details, constraint violations are penalized according to their distance to the satisfaction and integrated into the objective. As a matter of fact, it steers the generation towards satisfaction.
    
\end{itemize}
While these methods are effective for generating fluent text, they do not inherently guarantee that generated sequences satisfy external constraints, motivating the integration of CP techniques. Figure~\ref{fig:cpml} provides an overview of the feasibility challenges associated with different approaches.

\subsection{Masked Language Models (MLMs)}
Masked Language Models (MLMs), such as BERT \cite{BERT}, predict missing tokens in a text by considering both the left and right context. This bidirectional nature allows MLMs to capture deeper contextual relationships compared to autoregressive models. Typically, a portion of the input tokens is replaced by a special \texttt{[MASK]} token, and the model is trained to recover these tokens based on the surrounding context while maximizing:
\begin{equation}
\sum_{w_i \in M} \log P(w_i \mid \tilde{W}),
\end{equation}
where \(M\) is the set of masked positions in the modified sequence \(\tilde{W}\). This capability is central to our approach, as it enables the propagation of constraints over the entire sequence rather than incrementally, thereby overcoming a key limitation of autoregressive generation .

\begin{figure*}[t]
    \centering
    \resizebox{\textwidth}{2in}{
    \includegraphics[width=\linewidth]{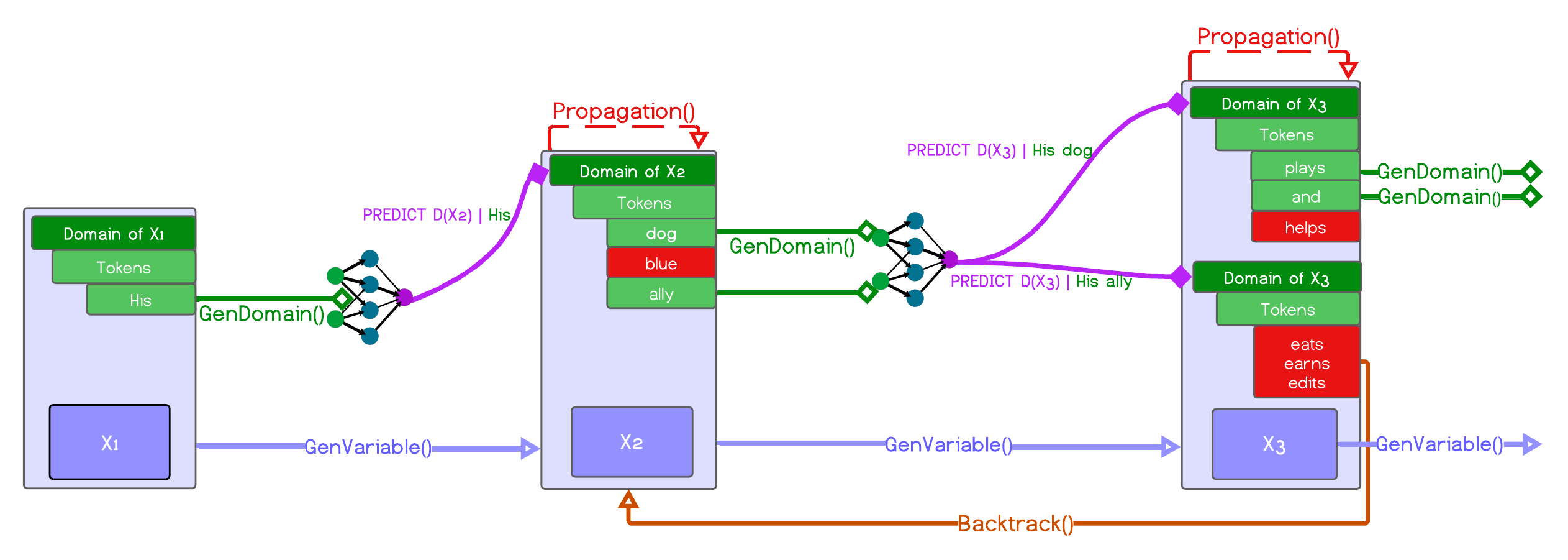}
    }
    \caption{Illustration of GenCP in a toy example where the task is to generate a passage without the letter "e". The autoregressive LLM incrementally predicts candidate domains for variables, while GenCP enforces constraints dynamically. If no valid domain remains for a variable, backtracking is triggered.}
    
    \label{fig:gencp}
    
\end{figure*}

\subsection{Constraint Satisfaction Problems (CSPs) and GenCP}
A Constraint Satisfaction Problem (CSP) is defined as a triplet \(\langle X, D, C \rangle\), where:
\begin{itemize}
    \item \(\mathcal{X} = \{X_1, X_2, \dots, X_n\}\) is the set of variables,
    \item \(\mathcal{D} = \{D(X_1), D(X_2), \dots, D(X_n)\}\) is the set of domains (each \(D(X_i)\) is the set of values the variable \(X_i\) can take), and
    \item \(\mathcal{C} = \{C_1, C_2, \dots, C_m\}\) is the set of constraints that specify allowable combinations of values.
\end{itemize}
A solution to a CSP is an assignment of values to all variables such that every constraint is satisfied.

Traditional CP approaches assume a fully defined CSP prior to the search. However, in constrained text generation, the structure of the problem is built incrementally as the text is generated. To address this, \textbf{GenCP}~\cite{regin-demaria:2023,regin-demaria-bonlarron:2024,regin:tel-04997072} introduces a dynamic, evolving CSP that is constructed on-the-fly. GenCP is governed by three core procedures:
\begin{itemize}
    \item \texttt{GenV}: Dynamically introduces new variables.
    \item \texttt{GenD}: Defines the domain of each new variable using LLM predictions.
    \item \texttt{GenC}: Imposes constraints over variables to enforce desired properties during generation.
\end{itemize}

In summary, GenCP starts with no variables, domains, or constraints and constructs them incrementally during the search for solutions. Whenever a domain is defined via \texttt{GenD}, classical CP techniques, such as constraint propagation, search, backtracking, and domain pruning, are employed to ensure that the generated text adheres to the specified constraints. Figure~\ref{fig:gencp} illustrates the entire process.


%% file: inc/method2.tex
\section{Future Variable Domain Generation with MLM}
\label{section:method}

In this section, we describe our proposed method for integrating bidirectional Masked Language Models (MLMs) into the dynamic CP framework GenCP for constrained text generation. Our approach enhances constraint propagation by leveraging MLM-based domain generation and is organized into three main components: variable representation (\texttt{GenV}), domain generation (\texttt{GenD}), and constraint integration (\texttt{GenC}).

\subsection{\texttt{GenV}: Variable Representation and Formatting}
\label{subsec:genv}
A key challenge in interfacing CP with language models is reconciling the granularity of decision variables with the token-based output of LLMs. While our formulation naturally models decision variables at the word level, LLMs operate on subword tokens. To bridge this gap, we introduce \emph{meta-variables} \(X_i\), where each meta-variable can comprise one or more decision variables \(X_{i_1}, X_{i_2}, \dots\).

For example, generating the word ``\texttt{Using}'' might involve:
\begin{itemize}
    \item \(X_{i_1}\): ``\texttt{Us}''
    \item \(X_{i_2}\): ``\texttt{ing}''
\end{itemize}
For instance, a current CSP assignment may be encoded as:
\[
\texttt{"Us;ing; a; transform;er;"}
\]
Here, spaces separate meta-variables, while semicolons distinguish individual decision variables. This flexible representation accommodates both token-level and word-level structure, and it can be extended to model larger textual units if needed.

\subsubsection{Variable Formatting}
As generation proceeds, most variables are already assigned. To generate the next variable, the MLM is provided with a prompt consisting of the current assignments followed by one or more \texttt{[MASK]} tokens. For example, if:
\[
X_1 = \texttt{"The"}, \quad X_2 = \texttt{"little"}, \quad X_3 = \texttt{"boy"},
\]
The prompt becomes:
\[
\texttt{"The little boy [MASK] [MASK] [MASK]\dots"}
\]
The MLM then predicts candidate tokens for the masked positions, forming the candidate domains \(D(X_4)\) and \(D(X_5)\). 

\subsection{\texttt{GenD}: Domain Generation via MLM}
\label{subsec:gend}
The \texttt{GenD} procedure is responsible for instantiating candidate domains for unassigned variables. In our framework, if a variable \(X_i\) is unassigned and masked, the system queries an MLM rather than an autoregressive LLM to infer its domain.

\subsubsection{MLM Domain Preview}
Given a partial assignment of \(k-1\) variables and a look-ahead depth \(d\), the MLM is used to generate candidate domains for the next \(d\) variables:
\[
\{ X_k,X_{k+1},...,X_{k+d}\}.
\]
After generating these candidate domains, relevant constraints are applied to the new variables. If any domain becomes empty, backtracking is triggered; otherwise, the next variable \(X_k\) is generated. Once the domain \(D(X_k)\) is obtained (potentially via an autoregressive LLM), it is filtered by propagators associated with the current set of constraints $\mathcal{C}$. Consequently, the assignment $X_k = w$, where $w \in D(X_k)$, is determined based on the remaining consistent values within all variable domains:
\[ \bigsqcup_{i=1}^{k+d}D(X_i). \]
This process can be iteratively applied to the next variable, $X_{k+1}$, considering now $k$ assignments.
Hence, it iteratively builds the complete text while ensuring that each assignment is consistent with past and anticipated future domains concerning the constraints.

In detail, following the generation of the domain for $X_k$ via an autoregressive LLM, the filtering of $D(X_k)$ is conducted using the propagators associated with the current set of constraints $C$.

\subsection{\texttt{GenC}: Constraint Integration and Propagation}
\label{subsec:genc}
After candidate domains are established, constraint propagation ensures that future assignments adhere to task-specific requirements. For example, a length constraint can be expressed as:
\[
C: \sum_{i=1}^{n} \#Char(X_i) = T,
\]
where \(T\) is the target total length and \(\#Char(X_i)\) denotes the number of characters in \(X_i\). Such format constraints are ubiquitous in constrained text generation tasks from poetry generation, summarization or text style transfer\cite{garbacea-arxiv-survey-nlp:2022}.

\subsubsection{Constraint-Specific Propagation}
Once \(k-1\) variables have been assigned, the solver prepares to assign the \(k^{th}\) variable by constructing a prompt:
\[
\{X_1 = w_1, \dots, X_{k-1} = w_{k-1}, X_k = [M], X_{k+1} = [M]\}.
\]
The MLM then predicts candidate values for \(X_{k+1}\), forming the domain \(D(X_{k+1})\). Constraint propagation refines \(D(X_{k+1})\) using the current assignments and the constraint \(C\). For instance, considering a length constraint, the solver computes bounds \(L\) and \(U\) such that:
\[
L \leq \#Char(X_{k+1}) \leq U,
\]
and adjusts the domains of current and future variables accordingly to ensure that the cumulative length satisfies constraint $C$.

\paragraph{Pathological Case.} In the worst-case, consider a sentence that must contain exactly ten words, with only two variables left to assign and ten characters remaining to satisfy the constraint \(C\). Standard knapsack filtering \cite{trick:03} cannot be applied directly because the domains of the last two variables are unknown. As a result, the search would explore all possible combinations in a brute-force, generate-and-test manner, iterating over \(D(X_{9}) \times D(X_{10})\) until it finds that, given \(D(X_{10})\), for instance it was (7+3).

MLMs help by eliminating inconsistent combinations early while prioritizing feasible ones. Instead of exhaustively exploring every token combination in the penultimate and final domains, the solver can precompute the Cartesian product of admissible token sums for these positions based on the preview gave by the MLM that satisfy the current sum constraint. Consequently, the solver considers only those tokens that are coherent with the constraint, enabling more fine assignments. This domain previewing reduces unnecessary exploration, and since the rankings of the combination preserve the likelihood ordering of the autoregressive LLM, the method does not interfere with the left-to-right generation process while integrating bidirectional information to improve constraint satisfaction and more efficiently reach feasible solutions.

\subsection{Implementation Workflow}
\label{subsec:workflow}
The complete workflow for integrating MLM-based domain generation into GenCP is as follows:
\begin{enumerate}
    \item \textbf{Domain Initialization:} Each variable \(X_i\) is initialized as empty or masked.
    \item \textbf{MLM Querying:} The MLM predicts candidate tokens for masked positions, forming the candidate domain \(D(X_i)\).
    \item \textbf{Constraint Propagation (filtering):} The CP solver refines \(D(X_i)\) based on previous assignments and task-specific constraints.
    \item \textbf{Decision Making (assignment):} A value is selected from \(D(X_i)\) to extend the partial assignment while ensuring consistency.
\end{enumerate}
This integration ensures that generated text satisfies both linguistic coherence and strict problem-specific constraints, addressing the limitations of purely autoregressive generation.


%% file: inc/result.tex
\section{Result}
\label{section:result}

\subsection{Experimental Conditions}
The experiments were performed on a laptop with Windows 10 Professional, 32 GB RAM, and Intel 16 CPU cores.
The new GenCP approach is implemented in Java 17. GenCP restarts the search each time a solution is found. A domain generation call (\texttt{GenD}) asks for 50 tokens (top-k) .

\subsubsection{Benchmarks}
Our benchmark suite consists of several task can be found in Tab. \ref{tab:task_constraints}.


These benchmarks are inspired by the COLLIE Benchmark \cite{yao2024collie}, which exposes the limitations of LLMs in handling such tasks alone. The task sent-1 is retained to exemplify a fundamental counting task, illustrating the advantage of MLM-based domain preview. The remaining tasks integrate keyword usage and counting within paragraph generation, a challenge not addressed in the original GenCP approach \cite{regin-demaria-bonlarron:2024}.

\begin{table}[ht]
    \centering
    \resizebox{!}{!}{ 
    \begin{tabular}{lrrl}
    \toprule
    Task & \#Sent & \#W p Sent & \#Char per Sent + Constraints  \\ 
    \midrule
    sent-1 & 1 &  & =82  \\ 
    para-2 & 2 & 10-15 & =60 \\
    para-3 & 3 & $\geq 15$ &  \\
    para-4 & 2 & 14 & 72-74  \\
    para-5 & 3 & & Start: "Dragons", "Kindgoms", "Barbarians" \\
    para-6 & 4 & & No occurrence of "the", "and", "of" \\
    \bottomrule
    \end{tabular}
    }
    \caption{One sentence generation task, (sent-1) and 5 paragraph task (para-2-6). : \#Sent = number of sentences, \#Words/\#Sent = word count per sentence, \#Char/\#Sent = character count per sentence with if any specific constraints.}
    \label{tab:task_constraints}
\end{table}

\begin{table}[ht]
    \centering
    \resizebox{!}{!}{ 
    \begin{tabular}{lccccc}
     \toprule
    sent-1    & d & \#LLMCalls & \#MLMCalls & \#bks & \#sols \\
    \midrule
    \texttt{vanilla}    & 0        & 444             & 0           & 375        & 3         \\
   \texttt{metavar}    & 0        & 456               & 0           & 211        & 4         \\
    \texttt{previewMLM} & 2        & \textbf{359}             & 32          & 170       & \textbf{10}        \\
    \midrule
    para-2 & \\
    \midrule
    \texttt{metavar} & 0 &446 & 0 & 136 & 3 \\
    \texttt{previewMLM}&	2&	\textbf{341}	&40&	193&	\textbf{6} \\

    \midrule
    para-3 & \\
    \midrule
   \texttt{metavar} &	0&	824	&0	&0	&13 \\
 
    \midrule
    para-4 & \\
    \midrule
   \texttt{metavar}        &0	&466&	0&	  447&	0 \\
   \texttt{previewMLM}	    & 2	&\textbf{422}&	12& 258&	\textbf{2} \\
 \midrule
 para-5 & \\
 \midrule
\texttt{metavar}&	0&	833&	0&	0&	21 \\

    \midrule
    para-6& \\
    \midrule
    \texttt{metavar} &	0	&430	& 0	& 34	&3 \\
    \midrule

    \end{tabular}
    }
    \caption{The table reports, for each task which GenCP version was used (e.g., \texttt{vanilla}, \texttt{metavar}, \texttt{previewMLM}, the depth $d$ for domain preview, the number of autoregressive LLM calls (\#LLMCalls), the number of MLM calls (\#MLMCalls), the number of backtracks (\#bks), and the number of solutions (\#sols) for sent-1 and paragraph tasks (para-2 to para-6).}
    \label{tab:overrallresult}
\end{table}

\begin{table}[h]
    \centering
    \resizebox{!}{!}{
    \begin{tabular}{lc}
    \toprule
    \textbf{task:} \\
    \textbf{generated sample...} \\
    \midrule

sent-1:\\

His eyes sparkled as he stared into the void, awaiting the arrival of an immortal.\\
\midrule
para-2: \\
With an arm raised in triumph, he raised himself once again. \\
The sun had shifted its position, casting a shadow on earth. \\ \midrule
para-3:\\
The warrior had been on the battlefield for at least a dozen years, \\
fighting and dying with the forces of the Empire \\ against the growing power of the Orcs. \\ 
He was the last in a line of warriors that had fought on the front. \\ 
The orcish king was dead, his son and grandson had been killed in a ruthless battle. \\
\midrule
para-4:\\
The warrior was an ancient warrior, known to many as The Sword of Justice. \\ 
His name is known to all, but he is known, mostly, throughout the lands. \\
\midrule
para-5:\\
Dragons raised into the sky, one of which bore a striking resemblance \\ to the dragons of myth,
were beginning to circle in the sky above. \\ Kingdoms trembled, while the mighty nations of the world sat abed, \\ 
awaiting the day's end. Barbarians with swords and axes drew together,
\\ their bloodlust and greed being satisfied one final time. \\
\midrule
para-6:\\
 He stood watching, waiting, as an eerie silence filled his mind. \\ His soul was trapped in a timeless battle, between \\ desire, truth, love, hate, fear, anger, lust, greed, envy, 
 \\ cowardice, pride, ambition, pride, lust, revenge, forgiveness, jealousy, rage, \\ 
 lust, hatred, ignorance, stupidity, ignorance, stupidity, etc. \\ 
 He was battling for his very soul. This was his life. \\

         \bottomrule
    \end{tabular}
    }
    \caption{One generated example with GenCP with \texttt{previewMLM} for the tasks 1,2 and 4. While, GenCP with \texttt{metavar} only for the tasks 3,5 and 6.}
    \label{tab:my_label}
\end{table}

\subsubsection{LLM and MLM Choice}
In the original GenCP paper, the model used was LLaMa Q4 7B, a lightweight variant of LLaMa quantized to 4-bit integers. The quantization level was such that it was unclear whether any loss of “meaning” was due to the search process or simply to poor token predictions. In contrast, in this paper we employ the autoregressive LLM \texttt{babbage-002} provided by OpenAI. This mid-sized model is not as large as the latest GPT-4 and can be considered as a basic model, lacking the additional refinements like InstructGPT. Thus, it is LLM that was not fine-tuned to follows instructions.

For the MLM component, computational considerations were less of an issue given that MLMs typically have fewer than 1 billion parameters. We used BERT as our MLM. Although more recent and larger MLM models are available, our objective is to gain insight into future values rather than to perfectly model language in this framework; therefore, employing a lightweight model is reasonable.

For left-to-right generation, we used \texttt{babbage-002}. The temperature was set to 0.8, it is considered as an high enough value for creative task (i.e., less generic output).
For domain preview, we used \texttt{bert-base-cased}. 


\paragraph{Pre-prompt:}
Directly prompting the LLM or the MLM with an empty variable assignments often results in dull and unengaging outputs. To address this, we introduced thematic grounding to inspire more vivid and compelling sentences than typical legal text or generic blogging. 
For instance, using the pre-prompt:\textit{“Amidst the crimson glow of a setting sun, a lone warrior, clad in battle-worn silver, stood atop the ancient ruins, his blade gleaming with the promise of legend.”} This pre-prompt was employed to guide the LLM towards generating related content, enhancing engagement without the primary intention of controlling semantics or themes. While the main focus of this article is not to manipulate thematically through pre-prompts while enforcing constraints, this technique demonstrates the potential to evoke heroic fantasy elements in the generated text.

\paragraph{Evaluation:}
Interacting with an LLM or an MLM often incurs high latency, typically on the order of hundreds of milliseconds. However, this is only a rough estimate since a wide range of hardware, from CPUs to GPUs, can run LLMs, and numerous optimization techniques (e.g., fast attention \cite{dao2022flashattention}, model quantization \cite{quantization_survey:2022}) can significantly accelerate inference times. To quantify our results, we evaluate our approach in terms of LLM calls. And it offers a form of hardware independence. As shown in Table~\ref{tab:overrallresult}, we use the number of LLM calls to provide insight into the interaction between GenCP and the LLMs, and to enable comparisons across experiments conducted under different settings.

\subsection{Result Analysis}

Our experimental results, summarized in Table~\ref{tab:overrallresult}, provide insight into both the efficiency and quality of the different GenCP variants across the benchmark tasks. In the following, we analyze the performance and output quality of the approaches in detail.

\paragraph{Quality Analysis:}
Across all benchmarks in Table~\ref{tab:task_constraints}, both the \texttt{vanilla} and \texttt{metavar} approaches find a similar number of solutions. However, when a word is split into several tokens using meta-variables, the performance in terms of constraint satisfaction can sometimes decrease. This is due to the increased number of combinations that must be considered to assign a suitable value to a single meta-variable representing a word. Nevertheless, the improvement in text quality and diversity with meta-variables is substantial. Without meta-variables, GenCP effectively limits its search to a small subset of the LLM's vocabulary, typically very short and frequent tokens (e.g., "as," "they," "their," "a," "them"). In contrast, incorporating meta-variables expands the search space to cover a broader vocabulary, which, although it increases the search space, ultimately leads to richer and more diverse generated text. As the next subsection will details, in case where the constraint is not local but applied over the whole sequence of text, the use of the \texttt{PreviewMLM} to find more solutions in the same amount of times.

\paragraph{Performance Analysis:}
For the single-sentence generation task (sent-1), the \texttt{vanilla} approach required 444 LLM calls, encountered 375 backtracks, and generated 3 valid solutions. The \texttt{metavar} variant slightly increased the number of LLM calls to 456 while reducing backtracks to 211, yielding 4 solutions. In contrast, the \texttt{previewMLM} method, which leverages a domain preview (with depth $d=2$) and incorporates an additional 32 MLM calls, significantly outperformed the other approaches by reducing LLM calls to 359 and backtracks to 170, while producing a notable 10 solutions. This clearly demonstrates that the domain preview strategy not only minimizes the reliance on expensive LLM calls but also improves the overall efficiency of the search.

A similar trend is observed in the paragraph-level tasks. In para-2, the \texttt{metavar} approach incurred 446 LLM calls and 136 backtracks to produce 3 solutions, whereas \texttt{previewMLM} reduced the LLM calls to 341 and, with 40 MLM calls and 193 backtracks, doubled the number of solutions to 6. For para-3, only the \texttt{metavar} variant was employed, requiring 824 LLM calls (with no backtracking) to yield 13 solutions. In para-4, the \texttt{metavar} method performed poorly with 466 LLM calls and a high backtracking cost of 447, failing to produce any valid solution; meanwhile, \texttt{previewMLM} achieved 422 LLM calls, 12 MLM calls, and 258 backtracks, resulting in 2 solutions. For para-5 and para-6, only the \texttt{metavar} method was used, with para-5 achieving 21 solutions from 833 LLM calls and para-6 yielding 3 solutions from 430 LLM calls and 34 backtracks.

These results highlight that integrating a domain preview (as in \texttt{previewMLM}) is particularly effective in reducing the number of LLM calls and reducing the search exploration, especially for tasks that impose strong constraints.

These observations suggest that while the non-preview approaches may be adequate for tasks with relatively weak constraints, the domain preview mechanism in \texttt{previewMLM} is particularly effective in more challenging scenarios. The additional guidance provided by the MLM preview seems to enable a more targeted search, thereby improving the feasibility capability of the approach to produces outputs under strict constraints.

\paragraph{Limitations:}
Since MLMs are trained to predict a limited fraction (typically about 15\%) of masked tokens, they perform best when given rich contextual information. Consequently, it is important to note that, the depth $d$ to define the preview of the domain is used only on the last two variables of any sentences. We tried further depth $d=3$ and $d=4$ and so on but unfortunately the number of solutions decrease for the same amount of running time while keeping the same quality. It seems the domain preview works best for the propagation of the sum over the last and before last variables to perform filtering and assignment.


%% file: inc/discussion.tex
\section{Discussion}
\label{section:discussion}

\subsection{Thinking Fast and Slow AI}
A fundamental challenge in constrained text generation is balancing intuitive language fluency with strict constraint satisfaction. This aligns with the dual-process theory described by Kahneman \cite{kahneman2011thinking}, which distinguishes between fast, intuitive reasoning (System 1) and slow, deliberate reasoning (System 2). Prior work \cite{Booch_Fabiano_Horesh_Kate_Lenchner_Linck_Loreggia_Murgesan_Mattei_Rossi_Srivastava_2021} has drawn parallels between this framework and AI systems, associating data-driven, heuristic-based approaches (e.g., LLMs) with System 1, while aligning structured reasoning methods (e.g., CP solvers) with System 2. 

In our framework, the domain generation process implemented via \texttt{GenD} using LLMs or MLMs act as System 1, quickly proposing candidate tokens based on heuristic evaluation. In contrast, the CP component, responsible for filtering and assigning future values, operates as System 2, carefully enforcing constraints and guiding the overall generation process. Here, the CP solver maintains the primary control loop and invokes the heuristic domain generation only when needed, ensuring that the generated text remains both fluent and compliant with the desired constraints.

This integration effectively bridges the gap between intuitive language generation and rigorous constraint enforcement, harnessing the strengths of both fast, heuristic reasoning and slow, deliberate decision-making.

\subsection{Meaningful Content Constraint}
Our approach leverages the LLM's expertise in producing meaningful text to manage the domains of the CSP. In this framework, the LLM serves two roles. First, it implicitly defines a “meaningful content constraint” by generating an initial set of candidate values for each decision variable, in a similar way as Lazy Arc Consistency \cite{lazy-ac:1996}. Second, it provides a good branching heuristic based on likelihood that guides the solver’s decisions.
In other words, assign the values (tokens) in the same ordering as traditional decoding strategies.
As a result, when a the domain of a variable is determined, the solver assigns values that not only satisfy the explicit constraints but also maintain high semantic quality, with backtracking employed to recover from inconsistent or low-likelihood assignments.

\subsection{Backtracking with LLM}
\subsubsection{Strengths of Incorporating Backtracking}
Backtracking, as a complete search strategy, is traditionally considered less efficient than incomplete methods; however, a backtracking search with good heuristic often brings with few backtracks a viable solution (e.g., for scheduling problems~\cite{baptiste2001constraint}).
This is the case for the experiments in task para-6 and sent-1.
This limited backtracking can capture high-quality, constrained text that a greedy procedure might overlook \cite{regin-demaria-bonlarron:2024}. The inherent completeness of backtracking provides an attractive guarantee: even when free text generation may lead to harmful or incoherent outputs, backtracking may be a great tool to enforce those constraints robustly. Furthermore, employing a restart-based strategy after finding the first solution: as demonstrated in creative applications \cite{sprockeels-vanroy:2024}, can yield both diversity and improved quality without sacrificing efficiency.

\subsubsection{Potential Dangers and Mitigation Strategies}
The complete nature of backtracking search can drive the LLM to explore low-likelihood regions of the search space. While this might increase the feasibility (higher satisfaction rates), it risks degrading overall text quality. In other words, a solution that meets all constraints may still be semantically or stylistically ``suboptimal'' if it resides in an unlikely region of the LLM's output distribution. This empirical observation underscores the importance of balancing constraint satisfaction with likelihood. Therefore, we advocate for two mitigation strategies: (1) initiating search restarts once a predefined number of acceptable solutions (e.g., \(k\) solutions) are found, and (2) continuously tracking the likelihood of the current assignment and backtracking when this likelihood falls below a set threshold. Such measures can help maintain the semantic quality of generated text while still benefiting from the robustness of backtracking.

%% file: inc/Conclusion.tex
\section{Conclusion}
\label{section:conclusion}
This paper addresses the limitations of autoregressive LLMs in constrained text generation tasks, thanks to an enhanced GenCP that integrates MLMs for domain previewing. Our approach balances token-level predictions (\texttt{metavar}) and structural constraint enforcement, resulting in more reliable constraint management during generation. Through our empirical evaluations on COLLIE benchmarks, we demonstrated that incorporating an MLM-based domain preview (\texttt{previewMLM}) significantly improves GenCP's performance in tasks with strict content constraints in terms of feasibility. Even though, GenCP uses additional MLM calls, it optimizes the use of LLM inferences, leading to more feasible and meaningful constrained text outputs.  By refining this interplay between CP and LLM decoding, we aim to broaden the potential for these systems to generate text when language fluency and constraint satisfaction are paramount.

\section*{Acknowledgments}
This work has been supported by the French government, through the 3IA C\^ote d'Azur Investments in the
Future project managed by the National Research Agency (ANR) with the reference number ANR-19-P3IA-0002.

%% file: main.bbl
\begin{thebibliography}{10}

\bibitem{yao2024collie}
Shunyu Yao, Howard Chen, Austin~W. Hanjie, Runzhe Yang, and Karthik~R Narasimhan.
\newblock {COLLIE}: Systematic construction of constrained text generation tasks.
\newblock In {\em The Twelfth International Conference on Learning Representations}, 2024.

\bibitem{cardei2025constraineddiscretediffusion}
Michael Cardei, Jacob~K Christopher, Thomas Hartvigsen, Brian~R. Bartoldson, Bhavya Kailkhura, and Ferdinando Fioretto.
\newblock Constrained discrete diffusion, 2025.

\bibitem{geh2025adversarialtokenization}
Renato~Lui Geh, Zilei Shao, and Guy~Van den Broeck.
\newblock Adversarial tokenization, 2025.

\bibitem{Chains-of-though:2022}
Jason Wei, Xuezhi Wang, Dale Schuurmans, Maarten Bosma, Brian Ichter, Fei Xia, Ed~H. Chi, Quoc~V. Le, and Denny Zhou.
\newblock Chain-of-thought prompting elicits reasoning in large language models.
\newblock In Sanmi Koyejo, S.~Mohamed, A.~Agarwal, Danielle Belgrave, K.~Cho, and A.~Oh, editors, {\em Advances in Neural Information Processing Systems 35: Annual Conference on Neural Information Processing Systems 2022, NeurIPS 2022, New Orleans, LA, USA, November 28 - December 9, 2022}, 2022.

\bibitem{Yao-tot:2023}
Shunyu Yao, Dian Yu, Jeffrey Zhao, Izhak Shafran, Thomas~L. Griffiths, Yuan Cao, and Karthik Narasimhan.
\newblock Tree of thoughts: deliberate problem solving with large language models.
\newblock In {\em Proceedings of the 37th International Conference on Neural Information Processing Systems}, NIPS '23, Red Hook, NY, USA, 2023. Curran Associates Inc.

\bibitem{beamsearch2:2018}
Matt Post and David Vilar.
\newblock Fast lexically constrained decoding with dynamic beam allocation for neural machine translation.
\newblock In {\em Proceedings of the 2018 Conference of the North {A}merican Chapter of the Association for Computational Linguistics: Human Language Technologies, Volume 1 (Long Papers)}, pages 1314--1324, New Orleans, Louisiana, June 2018. Association for Computational Linguistics.

\bibitem{lu-etal-2022-neurologic}
Ximing Lu, Sean Welleck, Peter West, Liwei Jiang, Jungo Kasai, Daniel Khashabi, Ronan Le~Bras, Lianhui Qin, Youngjae Yu, Rowan Zellers, Noah~A. Smith, and Yejin Choi.
\newblock {N}euro{L}ogic {A}*esque decoding: Constrained text generation with lookahead heuristics.
\newblock In {\em Proceedings of the 2022 Conference of the North American Chapter of the Association for Computational Linguistics: Human Language Technologies}, pages 780--799, Seattle, United States, July 2022. Association for Computational Linguistics.

\bibitem{bonlarron-et-al:2023}
Alexandre Bonlarron, Aurélie Calabrèse, Pierre Kornprobst, and Jean-Charles Régin.
\newblock Constraints first: A new mdd-based model to generate sentences under constraints.
\newblock In Edith Elkind, editor, {\em Proceedings of the Thirty-Second International Joint Conference on Artificial Intelligence, {IJCAI-23}}, pages 1893--1901, 2023.

\bibitem{bonlarron-regin:2024inter}
Alexandre Bonlarron and Jean-Charles Régin.
\newblock Intertwining cp and nlp: The generation of unreasonably constrained sentences.
\newblock In Kate Larson, editor, {\em Proceedings of the Thirty-Third International Joint Conference on Artificial Intelligence, {IJCAI-24}}, pages 7600--7608. International Joint Conferences on Artificial Intelligence Organization, 8 2024.
\newblock AI, Arts \& Creativity.

\bibitem{bonlarron-regin:2024markov}
Alexandre Bonlarron and Jean-Charles Régin.
\newblock Markov constraint as large language model surrogate.
\newblock In Kate Larson, editor, {\em Proceedings of the Thirty-Third International Joint Conference on Artificial Intelligence, {IJCAI-24}}, pages 1844--1852. International Joint Conferences on Artificial Intelligence Organization, 8 2024.
\newblock Main Track.

\bibitem{regin-demaria-bonlarron:2024}
Florian Régin, Elisabetta De~Maria, and Alexandre Bonlarron.
\newblock {Combining Constraint Programming Reasoning with Large Language Model Predictions}.
\newblock In {\em 30th International Conference on Principles and Practice of Constraint Programming (CP 2024)}, 2024.

\bibitem{top-p-ICLR-2020}
Ari Holtzman, Jan Buys, Li~Du, Maxwell Forbes, and Yejin Choi.
\newblock The curious case of neural text degeneration.
\newblock In {\em 8th International Conference on Learning Representations, {ICLR} 2020, Addis Ababa, Ethiopia, April 26-30, 2020}. OpenReview.net, 2020.

\bibitem{geh2024signal}
Renato Geh, Honghua Zhang, Kareem Ahmed, Benjie Wang, and Guy Van Den~Broeck.
\newblock Where is the signal in tokenization space?
\newblock In {\em Proceedings of the 2024 Conference on Empirical Methods in Natural Language Processing}, pages 3966--3979, 2024.

\bibitem{pachet-roy:2011}
Fran\c{c}ois Pachet and Pierre Roy.
\newblock Markov constraints: Steerable generation of markov sequences.
\newblock {\em Constraints}, 16(2):148–172, apr 2011.

\bibitem{BERT}
Jacob Devlin, Ming{-}Wei Chang, Kenton Lee, and Kristina Toutanova.
\newblock {BERT:} pre-training of deep bidirectional transformers for language understanding.
\newblock In Jill Burstein, Christy Doran, and Thamar Solorio, editors, {\em Proceedings of the 2019 Conference of the North American Chapter of the Association for Computational Linguistics: Human Language Technologies, {NAACL-HLT} 2019, Minneapolis, MN, USA, June 2-7, 2019, Volume 1 (Long and Short Papers)}, pages 4171--4186. Association for Computational Linguistics, 2019.

\bibitem{bessiere2006constraint}
Christian Bessiere.
\newblock Constraint propagation.
\newblock In {\em Foundations of Artificial Intelligence}, volume~2, pages 29--83. Elsevier, 2006.

\bibitem{barbieri-et-al-lyrics:2012}
Gabriele Barbieri, Fran\c{c}ois Pachet, Pierre Roy, and Mirko~Degli Esposti.
\newblock Markov constraints for generating lyrics with style.
\newblock In {\em Proceedings of the 20th European Conference on Artificial Intelligence}, ECAI'12, page 115–120, NLD, 2012. IOS Press.

\bibitem{papadopoulos-roy-pachet:2014}
Alexandre Papadopoulos, Pierre Roy, and Fran\c{c}ois Pachet.
\newblock Avoiding plagiarism in markov sequence generation.
\newblock In {\em Proceedings of the Twenty-Eighth AAAI Conference on Artificial Intelligence}, AAAI'14, page 2731–2737. AAAI Press, 2014.

\bibitem{papadopoulos-roy-etal:15}
Alexandre Papadopoulos, Pierre Roy, Jean-Charles R\'{e}gin, and François Pachet.
\newblock Generating all possible palindromes from ngram corpora.
\newblock In {\em Proceedings of the 24th International Conference on Artificial Intelligence}, IJCAI'15, page 2489–2495. AAAI Press, 2015.

\bibitem{papadopoulos-roy-pachet:2016assisted}
Alexandre Papadopoulos, Pierre Roy, and Fran{\c{c}}ois Pachet.
\newblock Assisted lead sheet composition using flowcomposer.
\newblock In Michel Rueher, editor, {\em Principles and Practice of Constraint Programming}, pages 769--785, Cham, 2016. Springer International Publishing.

\bibitem{sprockeels-vanroy:2024}
Damien Sprockeels and Peter Van~Roy.
\newblock Expressing musical ideas with constraint programming using a model of tonal harmony.
\newblock In Kate Larson, editor, {\em Proceedings of the Thirty-Third International Joint Conference on Artificial Intelligence, {IJCAI-24}}, pages 7753--7761. International Joint Conferences on Artificial Intelligence Organization, 8 2024.
\newblock AI, Arts \& Creativity.

\bibitem{hokamp-liu:2017}
Chris Hokamp and Qun Liu.
\newblock Lexically constrained decoding for sequence generation using grid beam search.
\newblock In {\em Proceedings of the 55th Annual Meeting of the Association for Computational Linguistics (Volume 1: Long Papers)}, pages 1535--1546, Vancouver, Canada, July 2017. Association for Computational Linguistics.

\bibitem{lu-etal-2021-neurologic}
Ximing Lu, Peter West, Rowan Zellers, Ronan Le~Bras, Chandra Bhagavatula, and Yejin Choi.
\newblock {N}euro{L}ogic decoding: (un)supervised neural text generation with predicate logic constraints.
\newblock In Kristina Toutanova, Anna Rumshisky, Luke Zettlemoyer, Dilek Hakkani-Tur, Iz~Beltagy, Steven Bethard, Ryan Cotterell, Tanmoy Chakraborty, and Yichao Zhou, editors, {\em Proceedings of the 2021 Conference of the North American Chapter of the Association for Computational Linguistics: Human Language Technologies}, pages 4288--4299, Online, June 2021. Association for Computational Linguistics.

\bibitem{regin-demaria:2023}
Florian Régin and Elisabetta De~Maria.
\newblock Using on-the-fly model checking to improve constraint programming for dynamic problems.
\newblock In {\em 2023 IEEE 35th International Conference on Tools with Artificial Intelligence (ICTAI)}, pages 393--398, 2023.

\bibitem{regin:tel-04997072}
Florian R{\'e}gin.
\newblock {\em {Generative constraint programming}}.
\newblock Theses, {Universit{\'e} C{\^o}te d'Azur}, December 2024.

\bibitem{garbacea-arxiv-survey-nlp:2022}
Cristina Garbacea and Qiaozhu Mei.
\newblock Why is constrained neural language generation particularly challenging?
\newblock {\em arXiv preprint arXiv:2206.05395}, 2022.

\bibitem{trick:03}
Michael Trick.
\newblock A dynamic programming approach for consistency and propagation for knapsack constraints.
\newblock {\em Annals of Operations Research}, 118:73–84, 2003.

\bibitem{dao2022flashattention}
Tri Dao, Dan Fu, Stefano Ermon, Atri Rudra, and Christopher R{\'e}.
\newblock Flashattention: Fast and memory-efficient exact attention with io-awareness.
\newblock {\em Advances in Neural Information Processing Systems}, 35:16344--16359, 2022.

\bibitem{quantization_survey:2022}
Amir Gholami, Sehoon Kim, Zhen Dong, Zhewei Yao, Michael~W. Mahoney, and Kurt Keutzer.
\newblock A survey of quantization methods for efficient neural network inference.
\newblock {\em CoRR}, abs/2103.13630, 2021.

\bibitem{kahneman2011thinking}
Daniel Kahneman.
\newblock {\em Thinking, fast and slow}.
\newblock Farrar, Straus and Giroux, New York, 2011.

\bibitem{Booch_Fabiano_Horesh_Kate_Lenchner_Linck_Loreggia_Murgesan_Mattei_Rossi_Srivastava_2021}
Grady Booch, Francesco Fabiano, Lior Horesh, Kiran Kate, Jonathan Lenchner, Nick Linck, Andreas Loreggia, Keerthiram Murgesan, Nicholas Mattei, Francesca Rossi, and Biplav Srivastava.
\newblock Thinking fast and slow in ai.
\newblock {\em Proceedings of the AAAI Conference on Artificial Intelligence}, 35(17):15042--15046, May 2021.

\bibitem{lazy-ac:1996}
Thomas Schiex, Jean-Charles Régin, Christine Gaspin, and Gerard Verfaillie.
\newblock Lazy arc consistency.
\newblock In {\em Proceedings of the AAAI Conference on Artificial Intelligence}, volume~1, pages 216--221, 1996.

\bibitem{baptiste2001constraint}
Philippe Baptiste, Claude Le~Pape, and Wim Nuijten.
\newblock {\em Constraint-based scheduling: applying constraint programming to scheduling problems}, volume~39.
\newblock Springer Science \& Business Media, 2001.

\end{thebibliography}
